\documentclass[10pt,twocolumn,letterpaper]{article}

\usepackage{cvpr}              

\usepackage{graphicx}
\usepackage{amsmath}
\usepackage{amssymb}
\usepackage{booktabs}
\usepackage{mathtools}
\usepackage{iac_pkg}
\usepackage{lipsum}
\usepackage{amssymb}
\usepackage{pifont}
\usepackage{multirow}
\usepackage{tabularx, booktabs}
\usepackage[htt]{hyphenat}
\usepackage{color}
\usepackage{xspace}
\usepackage{cite}
\usepackage{overpic}
\usepackage{arydshln}
\usepackage{subcaption}
\usepackage{xcolor}
\usepackage{bm}

\usepackage{cuted}

\definecolor{citecolor}{RGB}{34,139,34}
\definecolor{lightred}{RGB}{255,100,100}
\definecolor{cell_bisque}{rgb}{1.0, 0.89, 0.77}
\definecolor{cell_blond}{rgb}{0.98, 0.94, 0.75}
\definecolor{cell_blue}{RGB}{155, 187, 228}
\definecolor{princetonorange}{rgb}{1.0, 0.56, 0.0}
\definecolor{pinkpearl}{rgb}{0.91, 0.67, 0.81}
\definecolor{mossgreen}{rgb}{0.68, 0.87, 0.68}

\newcommand{\Paragraph}[1]{\vspace{-0mm}\noindent\textbf{#1.}\hspace{0mm}}
\newcommand{\Section}[1]{\vspace{-1mm} \section{#1} \vspace{-0mm}}
\newcommand{\SubSection}[1]{\vspace{-1mm} \subsection{#1} \vspace{-0mm}}

\usepackage{hyperref}

\hypersetup{pagebackref=true,breaklinks=true,letterpaper=true,colorlinks,bookmarks=false,citecolor=cyan}

\usepackage[capitalize]{cleveref}

\def\paperID{80} 
\def\confName{CVPR}
\def\confYear{2024}

\begin{document}
\title{DehazeDCT: Towards Effective Non-Homogeneous Dehazing via Deformable Convolutional Transformer}


\author{Wei Dong$^1$\quad Han Zhou$^1$\quad Ruiyi Wang$^2$\quad Xiaohong Liu$^{2}$\ \!\thanks{Corresponding author}\quad Guangtao Zhai$^2$\quad Jun Chen$^1$\\
$^1$McMaster University\quad
$^2$Shanghai Jiao Tong University\\
{\tt\small wdong1745376@gmail.com}, {\tt\small zhouh115@mcmaster.ca}\\ {\tt\small \{thomas25, xiaohongliu, zhaiguangtao\}@sjtu.edu.cn}, {\tt\small chenjun@mcmaster.ca}
}

\twocolumn[{
\maketitle
\vspace{-8mm}
\begin{center}
    \captionsetup{type=figure}
    \setlength{\abovecaptionskip}{2mm}
    \includegraphics[width=1\textwidth]{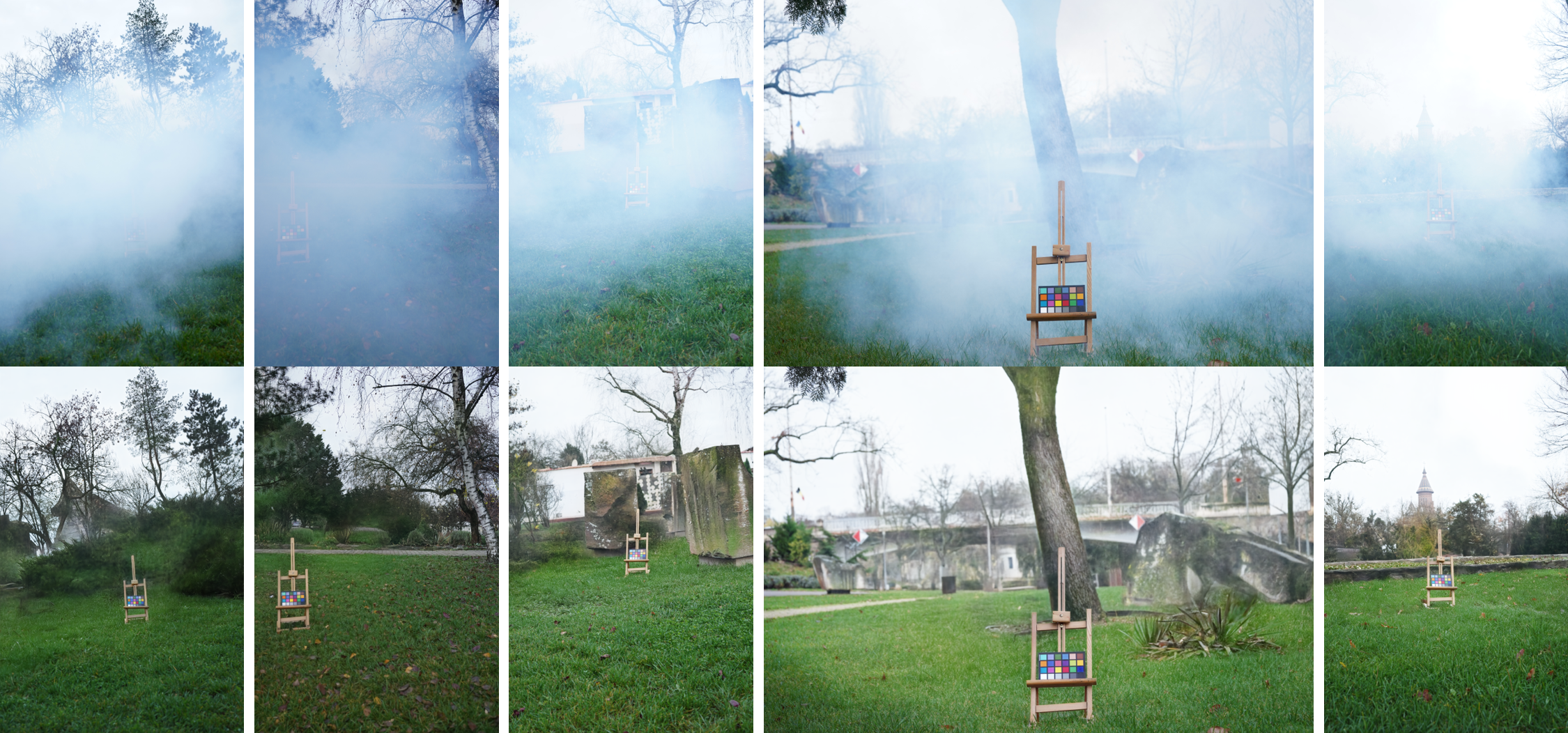}
    \captionof{figure}{Test Result of our method on NTIRE 2024 Dense and Non-Homogeneous Dehazing Challenge~\cite{NTIRE_Dehazing_2024}. Our \textbf{DehazeDCT} achieves the \textbf{second best} performance among 16 solutions and is capable to generate visually compelling outputs with vivid color and enhanced structure details.}
    \label{figure_challenge}
\end{center}
}]

\maketitle

\newcommand\blfootnote[1]{%
  \begingroup
  \renewcommand\thefootnote{}\footnote{#1}%
  \addtocounter{footnote}{-1}%
  \endgroup
}

\blfootnote{$^*$ Corresponding author}

\begin{abstract}
\vspace{-2mm}
Image dehazing, a pivotal task in low-level vision, aims to restore the visibility and detail from hazy images. Many deep learning methods with powerful representation learning capability demonstrate advanced performance on non-homogeneous dehazing, however, these methods usually struggle with processing high-resolution images (e.g., $4000 \times 6000$) due to their heavy computational demands. To address these challenges, we introduce an innovative non-homogeneous \textbf{Dehaz}ing m\textbf{e}thod via \textbf{D}eformable \textbf{C}onvolutional \textbf{T}ransformer-like architecture (\textbf{DehazeDCT}). Specifically, we first design a transformer-like network based on deformable convolution v4, which offers long-range dependency and adaptive spatial aggregation capabilities and demonstrates faster convergence and forward speed. Furthermore, we leverage a lightweight Retinex-inspired transformer to achieve color correction and structure refinement. Extensive experiment results and highly competitive performance of our method in NTIRE 2024 Dense and Non-Homogeneous Dehazing Challenge, ranking second among all 16 submissions, demonstrate the superior capability of our proposed method. The code is available: \url{https://github.com/movingforward100/Dehazing_R}.  
\end{abstract}

\section{Introduction}
\vspace{-1mm}
\label{sec:intro}
Images captured in hazy conditions, whether naturally occurring or artificially synthesized, share similar properties of low visibility, decreased contrast, and degraded structural details. These deteriorative characteristics severely impair the performance of various vision tasks, such as object recognition, tracking, and segmentation systems~\cite{tracking, segmentation, segmentation2, object-detection}, thereby hindering their application in hazy situations. This predicament urgently calls for dehazing methods that are effective across various scenarios.

Initiated upon the foundational principles of the atmospheric scattering model (ASM)~\cite{ASM}, early endeavors in image restoration~\cite{scatterdehaze1,scatterdehaze2,FMSNet} have been primarily oriented towards delineating the correlation between hazy images and their haze-free counterparts. The ASM can be succinctly articulated as follows: 
\begin{equation}
\setlength\abovedisplayskip{2pt}
\setlength\belowdisplayskip{2pt}
  I(x) = J(x)t(x) + A(1-t(x)),
  \label{eq:ASM}
\end{equation}
where $I$ and $J$ signify the hazy image and its clean counterpart; $x$ and $A$ represents the pixel location and the global atmosphere light; $t(x)$ denotes the transmission map , which is a function of the atmosphere scattering parameter $\beta$ and the scene depth, articulated as $t(x)=e^{-\beta d(x)}$. This formulation posits the image dehazing to the precise estimation of the transmission map $t(x)$ and the global atmosphere light $A$. Notwithstanding, the ASM presupposes an idealized uniform haze distribution, a limitation rendering the model less effective in addressing non-homogeneous dehazing challenges.

Recent advancements in image dehazing have been significantly influenced by the application of deep learning techniques~\cite{14,38,43,ACERNet,frequencydehazing2022, GridDehazeNet, GridDehazeNet+}, a development prompted by their profound success in areas such as classification and object detection. Notably, compared to the previously dominant ASM framework, deep learning-based methods have demonstrated superior performance on removing the haze from images with complex and spatially varying haze distributions, emerging as the predominant approach for tackling non-homogeneous dehazing problems.

Recent research has predominantly focused on exploring robust and powerful representation learning mechanisms to delineate the mappings between hazy and haze-free images. These methods usually learn spatial and frequency representations simultaneously, integrate special architectures to increase the receptive field, or utilize large scale CNN networks pre-trained on large datasets to harness transfer learning benefits. For example, DWT-FFC~\cite{DWT-FFC_2023_CVPRW} entails Discrete Wavelet Transform to capture spatial and spectral information effectively, employs Fast Fourier Convolution (FFC)~\cite{FFC,lama} to extend the receptive capacity, and harnesses the pre-trained ConvNext model to facilitate transfer learning; DehazeFormer~\cite{transformer_dehazing} introduces a transformer-based architecture with a shifted window partitioning scheme based on reflection padding for dehazing.

Nonetheless, current methods encounter certain challenges that necessitate further exploration: \textbf{First}, traditional Convolutional Neural Networks (CNNs) often suffer from strict inductive bias, and CNN-based models frequently rely on sizable fixed dense kernels (e.g., $31\times31$)~\cite{ding2022replknet,liu2022slak} to facilitate robust representation learning. This strategy not only incurs significant computational loads but also lacks the capacity for adaptive spatial aggregation conditioned by the input. \textbf{Second}, while transformer-based architectures are capable to capture long-range dependencies and facilitate adaptive spatial aggregation, they are hampered by computational and memory inefficiencies. The inherent complexity of the self-attention mechanism, which scales quadratically with input resolution, precludes their application in high-resolution dehazing scenarios, exemplified the $4000\times6000$ resolution image in DNH-HAZE dataset~\cite{NTIRE_Dehazing_2024}.

To tackle these challenges, we introduce \textbf{DehazeDCT}, a novel non-homogeneous \textbf{Dehaz}ing m\textbf{e}thod via \textbf{D}eformable \textbf{C}onvolutional \textbf{T}ransformer architecture. This model is comprised of two primary components: a dehazing module and a refinement module. In the \textbf{Dehazing} module, we engineer transformer-like dehazing branch that incorporates multiple DCNFormer blocks. Diverging from traditional Transformer architecture, our DCNFormer blocks utilize Deformable Convolution v4~\cite{xiong2024efficient} in lieu of the standard self-attention mechanism, thus ensuring our model benefits from both long-range dependency and adaptive spatial aggregation capabilities. Furthermore, by removing the redundant operation (softmax normalization in spatial aggregation) in traditional DCN~\cite{DCNv3}, our DCNFormer architecture demonstrates faster convergence and forward speed. Besides, we integrate a frequency-aware branch to facilitate the acquisition of frequency representations. In the \textbf{Refinement} module, we enact a lightweight Retinex-inspired transformer for color correction and structure refinement. With these dedicated designs, our \textbf{DehazeDCT} is capable to achieve color and detail consistency, essential for visually compelling results as shown in Fig.~\ref{figure_challenge}.

Our mian contributions are as follows:

$\diamond$ We propose an effective non-homogeneous dehazing method based on deformable convolutional transformer, followed by a Retinex-based transformer network for color and detail refinement.

$\diamond$ We design a DCNv4 based transformer-like network for dehazing, which offers long-range dependency and adaptive spatial aggregation capabilities and demonstrates faster convergence and forward speed.

$\diamond$ The effectiveness and generalization of our proposed method are verified by comprehensive experiment results on several benchmark datasets and the visually compelling performance in the NTIRE 2024 Dense and Non-Homogeneous Dehazing Challenge.
\begin{figure*}[!t]
    \setlength{\abovecaptionskip}{2mm}
    \centering
    \begin{overpic}[width=0.99\textwidth]{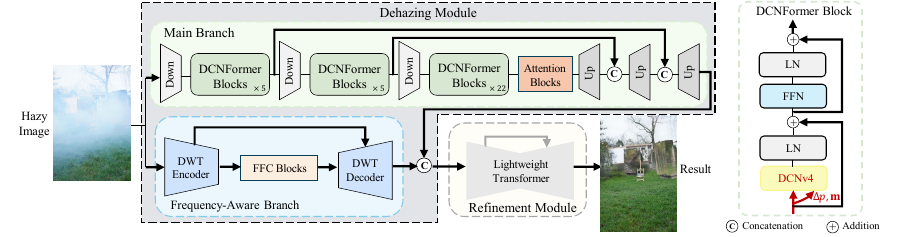}
    \end{overpic}
    \caption{ The overall architecture of our proposed model. In the Dehazing module, we introduce a transformer-like dehazing branch based on deformable convolution (DCNv4~\cite{xiong2024efficient}). In each DCNFormer block, DCNv4 is utilized to calculate the offset ($\Delta p $) and modulation scalar ($\mbf{m}$). Besides, the frequency-aware branch proposed in~\cite{DWT-FFC_2023_CVPRW} is also adopted as an auxiliary branch. In the Refinement module, we leverage a lightweight retinex-inspired transformer network to further reduce the color deviation and enhance texture details.} 
    \vspace{0mm}
    \label{fig_frame}
\end{figure*}

\section{Related Work}
\label{sec:related}

\Paragraph{Frequency Based Image Restoration} Similar to the spatial domain, the frequency domain encapsulates abundant information of images. In recent years, growing research attention has been drawn to take advantage of frequency information for image restoration~\cite{sr4, freq_yoo, freq_jiang, freq_yang, freq1}. Specifically, Yoo~\etal~\cite{freq_yoo} estimate the DCT coefficient for each frequency band for JPEG compression artifact removal. Yang~\etal~\cite{freq_yang} propose a wavelet U-Net where the down-sampling and up-sampling are replaced by discrete wavelet transform and its inverse operation for better reconstruction of edges and colors. Jiang~\etal~\cite{freq_jiang} introduce a novel focal frequency loss considering both amplitude and phase information and a dynamic spectrum weighting to adaptively guide existing models for frequency domain reconstruction. Cui~\etal~\cite{freq1} design a multi-branch dynamic selective frequency module (MDSF) which dynamically decomposes features into several frequency bands and uses channel-wise attention to highlight the useful frequency for image restoration.

\Paragraph{Image Restoration with Deformable Convolution} To overcome the limitation of traditional CNNs in modeling large and unknown transformations, deformable convolution~\cite{DCN} is proposed. By adding offsets to the regular sampling positions, deformable convolution allows flexible form of the sampling grid. Deformable convolutional networks have achieved remarkable performance on various high-level vision tasks~\cite{DCNv2, DCNv3, xiong2024efficient}. Recently, several works have devoted to apply deformable convolution for image restoration. Wang~\etal~\cite{deformable_restore} design a pyramid, cascading and deformable (PCD) alignment module, where deformable convolution is applied for feature-level frame alignment, for video restoration.  Wu~\etal~\cite{ACERNet} deploy two deformable convolutional layers after the deep layers of the autoencoder-like network for image dehazing. Wang~\etal~\cite{deformable_sr} propose an angular deformable alignment module (ADAM) where the deformable convolution is used to align features by their corresponding offsets.

\Paragraph{Transformer Based Image Dehazing} Due to the flexibility of capturing global dependencies, transformers have succeeded in various computer vision tasks. For instance, Uformer~\cite{Uformer}, SwinIR~\cite{swinir} and Restormer~\cite{Restormer} achieve remarkable performance on many image restoration tasks. Recently, multiple methods have applied transformers to single image dehazing. Based on Swin Transformer~\cite{liu2021Swin}, Song~\etal~\cite{transformer_dehazing} propose DehazeFormer with modified normalization layer, activation function and spatial information aggregation method. Guo~\etal~\cite{guo2022dehamer} combine transformer and CNN for image dehazing through transmission-aware 3D position embedding and feature modulation. Qiu~\etal~\cite{Qiu_2023_taylorformer} apply Taylor expansion to approximate the conventional softmax attention in transformer, which achieved linear complexity while retained the flexibility. These works demonstrate state-of-the-art performance on various dehazing benchmarks and inspired us to incorporate transformer into our model.
\begin{table*}[h]
\setlength{\abovecaptionskip}{2mm}
\centering

        \scalebox{1}{
        \begin{tabular}{c|cc|cc|cc|cc}
        \hline
        \multirow{2}{*}{\begin{tabular}{c}
            \textbf{ Methods}
        \end{tabular}}&\multicolumn{2}{c|}{NH-HAZE~\cite{NH-Haze_2020}} &\multicolumn{2}{c|}{NH-HAZE2~\cite{NTIRE_Dehazing_2021}} &\multicolumn{2}{c|}{HD-NH-HAZE~\cite{NTIRE_Dehazing_2023}} &\multicolumn{2}{c}{HD-NH-HAZE2~\cite{NTIRE_Dehazing_2024}} \\\cline{2-9}
        &PSNR $\uparrow$ &SSIM$\uparrow$  &PSNR$\uparrow$ &SSIM$\uparrow$ &PSNR$\uparrow$ &SSIM$\uparrow$   &PSNR$\uparrow$ &SSIM$\uparrow$\\ 
        \hline

        FFA~\cite{ffanet}                           &19.50 &0.644 &20.56 &0.811 &20.23 &0.710 &20.14 &0.707  \\
        TDN~\cite{TDN}                           &20.73 &0.673 &20.44 &0.801 &20.06 &0.713 &19.88 &0.700  \\
        AECR-Net~\cite{ACERNet}                      &19.88 &0.717 &20.75 &0.831 &20.34 &0.731 &20.26 &0.724  \\
        DWT-FFC~\cite{DWT-FFC_2023_CVPRW}                       &\blue{22.64} &0.730 &\blue{22.82} &\blue{0.874} &\blue{22.20} &\blue{0.746} &\blue{21.58} &\blue{0.738}  \\
        DehazeFormer~\cite{transformer_dehazing}                  &20.47 &\blue{0.731} &21.07 &0.825 &20.89 &0.728 &20.29 &0.718  \\
       
        \hline
        \textbf{DehazeDCT (Ours)} &\textbf{\red{22.78}} &\textbf{\red{0.734}} &\textbf{\red{22.86}} &\textbf{\red{0.877}} &\textbf{\red{22.36}} &\textbf{\red{0.752}} &\textbf{\red{21.73}} &\textbf{\red{0.743}} \\
        \hline
        \end{tabular}
    }
    
\caption{Quantitative comparisons between our proposed DehazeDCT and SOTA methods. Our proposed method achieves superior performance in terms of PSNR and SSIM across four datasets. These numbers are obtained from their original paper or training with their released code. [Key: \textbf{\red{Best}}, \blue{Second Best}, $\uparrow$ ($\downarrow$): Larger (smaller) values leads to better performance, HD-NH-HAZE2: Official dataset for NTIRE 2024 Dense and Non-Homogeneous Dehazing Challenge]}
\label{table_quant_compar}
\end{table*}

\section{Methods}
\label{sec:method}

The key contribution of our work is to design a novel dehazing module based on DCNv4~\cite{xiong2024efficient} (Deformable Convolution v4) and adopt a lightweight transformer for color and detail enhancement (Refinement module). The overview of our framework is provided as Fig.~\ref{fig_frame}, where the training process can be divided into two stages. In stage I, we optimize the Dehazing module to achieve preliminary dehazing (Sec.~\ref{sec_dehazing_module}, Sec.~\ref{sec_loss}). In stage II, the Refinement module is incorporated into the optimization process for detailed refinement. (Sec.~\ref{sec_refinement_module}).

\subsection{DCNv4 based Transformer for Dehazing}
\label{sec_dehazing_module}

Inspired by impressive ability of capturing long-range dependencies and facilitating adaptive spatial aggregation, we design a transformer-like branch, together with a frequency-aware branch similar to~\cite{DWT-FFC_2023_CVPRW, wei2024shadow}, for effective dehazing, as shown in Fig.~\ref{fig_frame}. Specifically, an hazy input $ \mbf{I} \in \mathbb{R}^{W \PLH H \PLH 3}$ is encoded into $\mbf{F}_{i} \in \mathbb{R}^{\frac{W}{4*2^{i}} \PLH \frac{W}{4*2^{i}} \PLH d_i}$ by $i$ downsampling operations, where $W$, $H$, and $d_i$ represent the image width, image height, the dimension of latent features. After each downsampling process, several DCNFormer blocks, which share similar architecture with common transformer blocks, are adopted for representation learning. However, instead of the global attention mechanism of transformers, the core operator of our DCNFormer is the deformable convolution v4, which is achieved by removing the softmax normalization operation in DCNv3~\cite{DCNv3}.

Given each latent feature $\mbf{F}_{i}$ and current pixel $p_0$, the principle of DCNv3 operation is described as:
\begin{equation}
    \textbf{y}(p_0) = \sum^{G}_{g=1} \sum^{K}_{k=1} \mathbf{w}_g \mathbf{m}_{gk}\mathbf{x}_g(p_0 + p_k + \Delta p_{gk}),
    \label{eqn_dcnv3}
\end{equation}
where $K$ represents the total number of sampling points, $k$ enumerates the sampling point and $G$ denotes the total number of aggregation groups.
For the $g$-th group,
$\mathbf{w}_g\!\in\!\mathbb{R}^{d_i\times d'_i}$ denotes the location-irrelevant projection weights of the group, where $d'_i\!=\!d_i/G$ represents the group dimension.
$\mathbf{m}_{gk}\!\in\!\mathbb{R}$ denotes the modulation scalar of the $k$-th sampling point in the $g$-th group, normalized by the softmax function along the dimension $K$.
$\mathbf{x}_g$ represents the sliced input feature map.
$\Delta p_{gk}$ is the offset corresponding to the grid sampling location $p_k$ in the $g$-the group.

Unlike utilizing softmax operation to normalize the modulation scalar $\mathbf{m}_{gk}$, we don't incorporate any normalization functions in our DCNFormer in order to achieve unbounded dynamic weights for $\mathbf{m}_{gk}$, which contributes to the significantly faster converge and forward speed compared to DCNV3~\cite{DCNv3}, common convolutions, and attention blocks in Transformers. Moreover, the core operator in our DCNFormer only adopts a $3 \times 3$ kernel to learn long-range dependencies, which is easier to be optimized compared to large kernels~\cite{ding2022replknet,liu2022slak}. Please note despite the comparatively large parameters of our dehazing module, our DCNFormer blocks allow our model to achieve high-resolution dehazing efficiently, without the need for special design of high-resolution images (e.g., $4000\times 6000$).

\subsection{Loss Function}
\label{sec_loss}

The loss function utilized for optimizing our Dehaizng module is:
\begin{equation}
 {L}_{loss}={L}_{1} + {\alpha}{L}_{SSIM}+{\beta}{L}_{Percep}+{\gamma}{L}_{adv},
\label{eq_loss} 
\end{equation}
where ${L}_{1}$, ${L}_{SSIM}$ and ${L}_{Percep}$ represent $L1$ loss, MS-SSIM loss ~\cite{DWT-FFC_2023_CVPRW}, and perceptual loss ~\cite{VGG-16}, respectively. In addition, we adopt the discriminator in ~\cite{GAN} to calculate adversarial loss (${L}_{adv}$). $\alpha$, $\beta$, and $\gamma$ are hyper-parameters and are set to $0.4$, $0.01$, and $0.0005$, respectively.

\subsection{Transformer based Refinement}
\label{sec_refinement_module}

Based our Dehazing module, which concentrates on removing haze from degraded images, we further incorporate a lightweight transformer similar to~\cite{retinexformer, AttentionLut} for refinement. Based on Retinex theory, given the input and its mean values for each pixel along the channel dimension, our Refinement module first predicts the lit-up image and light-up feature and then restore the color and detail corruption. Therefore, the our final dehazed image can be obtained by: 
\begin{equation}
\mbf{I}_{dehazed} = \bm{\phi} (\bm{\theta}(\mbf{I}_{hazy}), mean(\bm{\theta}(\mbf{I}_{hazy}))),
\label{eq_model}
\end{equation}
where $\bm{\theta}$ and $\bm{\phi}$ represent our Dehazing module and Refinement module, respectively. Notably, our Refinement module only has 1.61 M parameters, and we remove the adversarial loss in Eq.~\ref{eq_loss} to optimize the Refinement module $\bm{\phi}$.

\Section{Experiments}
\label{sec_ex}
\begin{figure*}[h]
    \setlength{\abovecaptionskip}{2mm}
    \centering
    \includegraphics[scale=0.68]{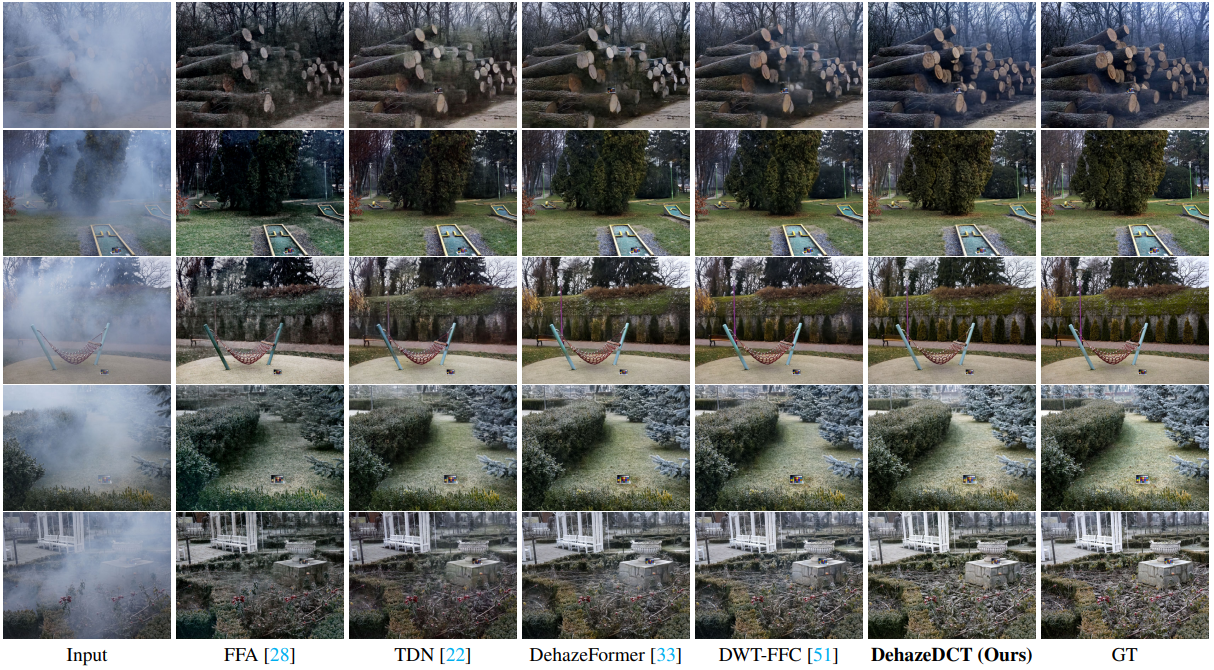}
    \caption{Visual comparisons on NH-HAZE~\cite{NH-Haze_2020} dataset. Compared to other models, our method exhibits higher color fidelity and effective dehazing, yielding compelling results. }
    \label{ntire20}
\end{figure*}

\subsection{Experiment Settings}
\Paragraph{Datasets} We qualitatively and quantitatively evaluate our proposed method on four real-world datasets: NH-HAZE~\cite{NH-Haze_2020}, NH-HAZE2~\cite{NTIRE_Dehazing_2021}, HD-NH-HAZE~\cite{NTIRE_Dehazing_2023} and DNH-HAZE2~\cite{NTIRE_Dehazing_2024} datasets. NH-HAZE dataset composes of 55 pairs of 1200$\times$1600 hazy and corresponding clean images. We use the official testing data for evaluation while the remaining are utilized for training. NH-HAZE2 dataset consists of 25 training pairs, 5 validation pairs and 5 testing pairs with resolution 1200$\times$1600. As the ground truth for validation and testing data isn't publicly accessible, we utilize the first 20 training pairs for training and use the rest as testing samples. HD-NH-HAZE and DNH-HAZE datasets compose of 40 training pairs, 5 validation pairs and 5 testing pairs of 4000$\times$6000 images. As we can only access the ground truth of training pairs, we again evaluate on the last 5 training pairs and use the rest for training.

\begin{figure*}[t]
    \setlength{\abovecaptionskip}{2mm}
    \centering
    \includegraphics[scale=0.67]{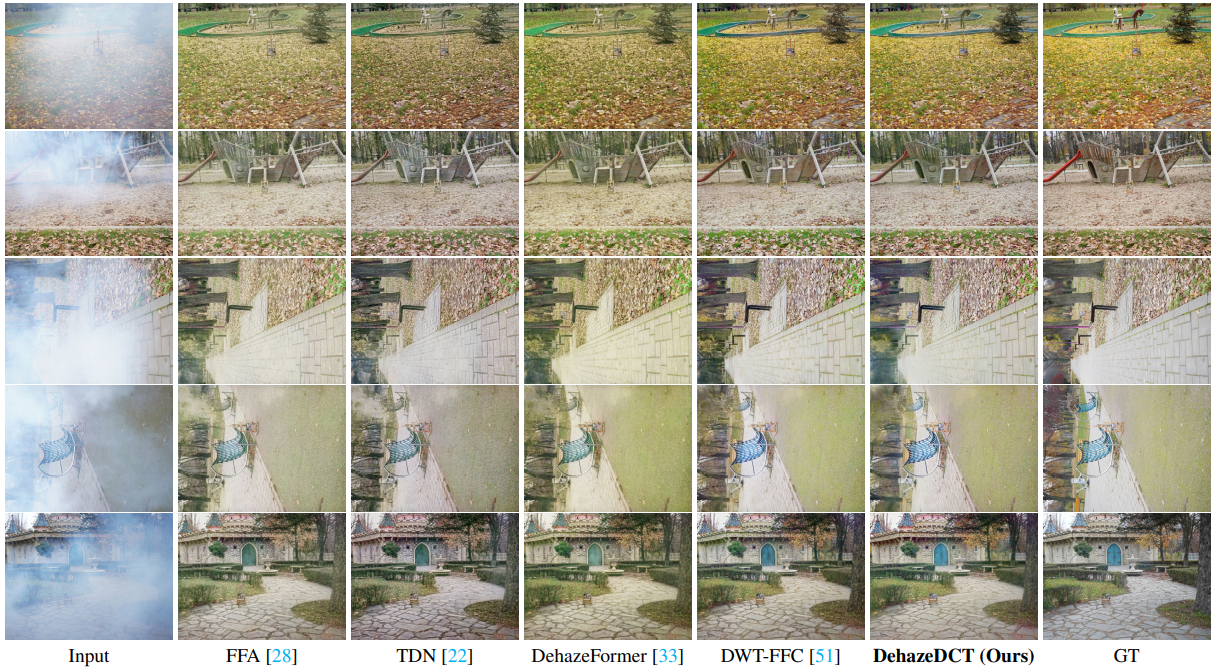}

    \caption{Visual experiment results on NH-HAZE~\cite{NTIRE_Dehazing_2021} dataset. Obviously, our method demonstrates superior performance on color preservation and detail maintaining, further enhancing the overall quality of the output.}
    \label{ntire21}
\end{figure*}
\begin{figure*}[ht]
    \setlength{\abovecaptionskip}{2mm}
    \centering
    \includegraphics[scale=0.92]{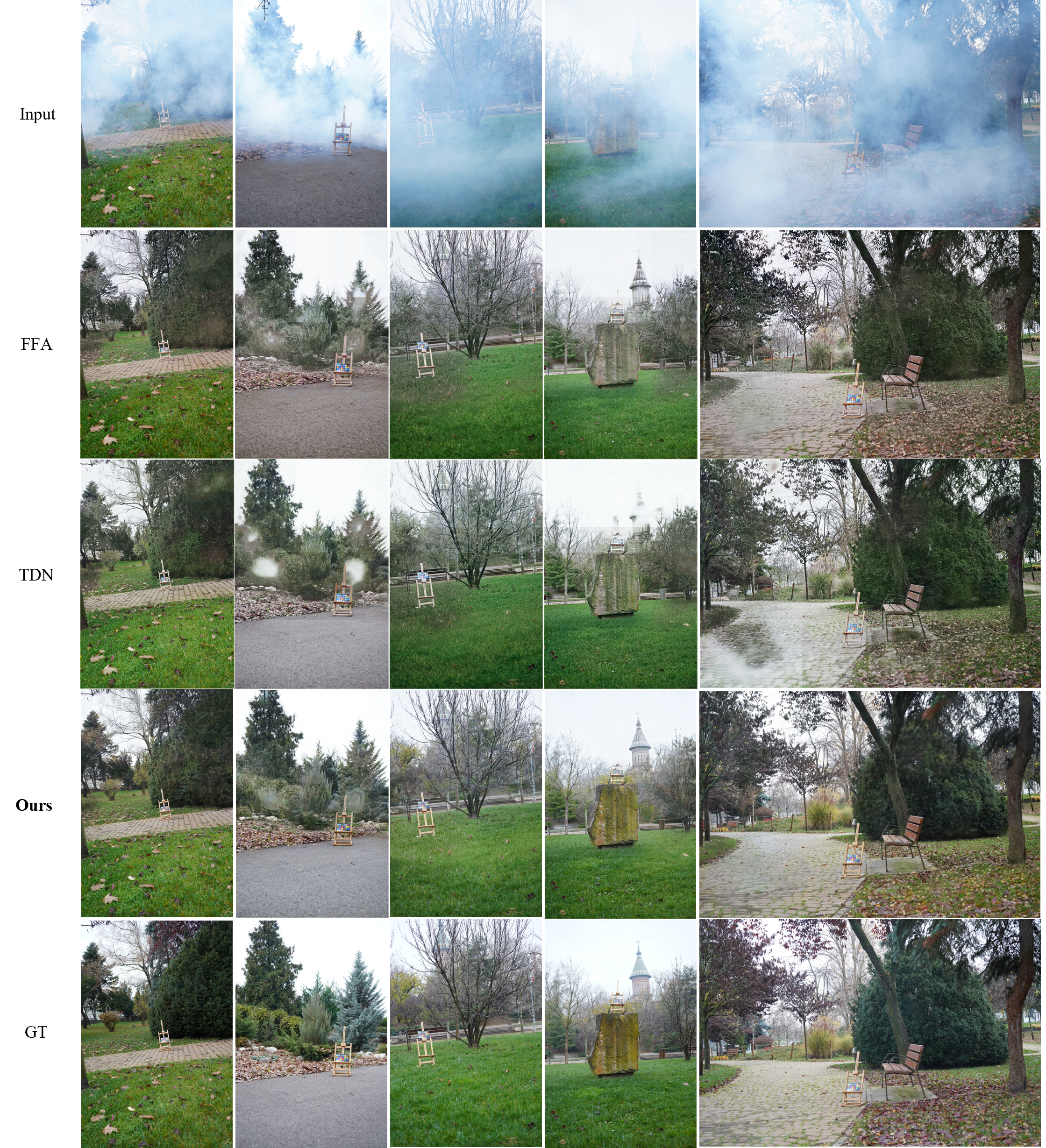}
    \caption{Visual Comparisons on HD-NH-HAZE dataset~\cite{NTIRE_Dehazing_2023}. Our method exhibits superior haze removal, evidenced by more vivid colors and clearer details, especially in the foliage and background structures.
} 
    \label{ntire23}
\end{figure*}

\begin{figure*}[ht]
    \setlength{\abovecaptionskip}{2mm}
    \centering
    \includegraphics[scale=0.96]{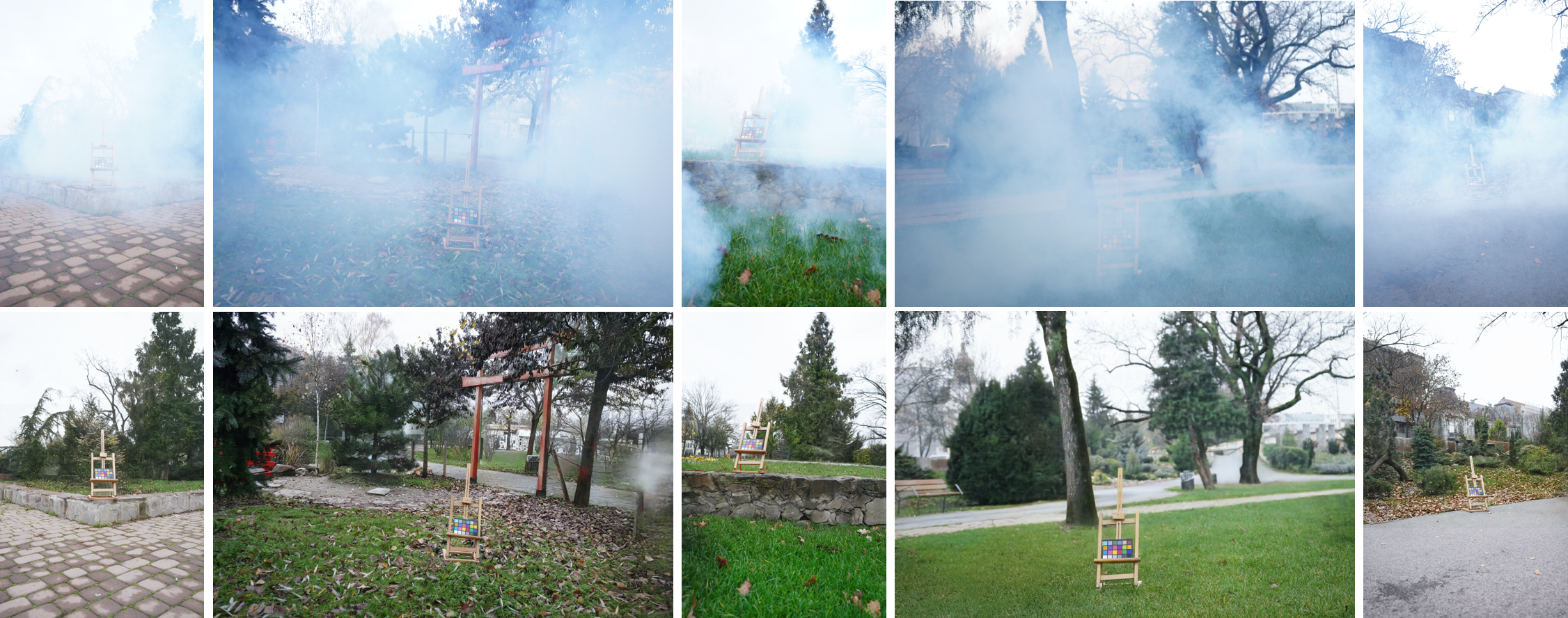}
    \caption{Our results on the validation set of NTIRE 2024 Dense and Non-Homogeneous Dehazing Challenge~\cite{NTIRE_Dehazing_2024}, achieving the best performance in terms of both PSNR and SSIM on the validation leaderboard.} 
    \label{fig_valid}
\end{figure*}


\Paragraph{Implementation Details} We implement the training using Pytorch 1.11.0 on an NVIDIA RTX 4090 GPU. To augment the limited training data, images pairs are randomly cropped into patches of size 384$\times$384, then possibly rotated at 90, 180, or 270 degrees, vertically or horizontally flipped. The training process of our proposed method composes of two stages. In stage I, we neglect the Refinement module. The model and discriminator are updated by Adam optimizer with decay factors $\beta_1=0.9$ and $\beta_2=0.999$ for a total of 5,000 epochs. The initial learning rate is set to be $1\times 10^{-4}$ and the learning rate decays by half at 1,500, 3,000, and 4,000 epochs. In stage II, we train our the Refinement module for 500 epochs and optimize our whole model for 500 epoch with the fixed the learning rate of $1\times 10^{-5}$.

\subsection{Comparisons with SOTA Methods}
\Paragraph{Compared Methods and Evaluation Metrics} In this section, we undertake a comprehensive evaluation of our method by quantitatively and qualitatively comparing it with current SOTA methods for dehazing. These benchmark models include the winner solution of NTIRE 2020 Non-Homogeneous Dehazing Challenge (TDN~\cite{TDN, NTIRE_Dehazing_2020}), the champion method in NTIRE 2023 HR Non-Homogeneous Dehazing Challenge (DWT-FFC~\cite{DWT-FFC_2023_CVPRW}), as well as FFA~\cite{ffanet}, AECR-Net~\cite{ACERNet}, and the recently proposed transformer-based dehazing approach (DehazeFormer~\cite{transformer_dehazing}). Besides, we utilize two full-reference metrics for quantitative evaluation: Peak Signal-to-Noise Ratio (PSNR) and Structural Similarity Index Measure (SSIM~\cite{SSIM}), which measures the pixel-level accuracy and the structural similarity of dehazed results.

\Paragraph{Quantitative Comparisons}
Tab.~\ref{table_quant_compar} presents the performance comparisons between our DehazeDCT and other methods across various datasets. As illustrated in Tab.~\ref{table_quant_compar}, our DehazeDCT consistently demonstrates superior performance in terms of both PSNR and SSIM across all four datasets, with 21.73 dB PSNR and 0.743 SSIM on DNH-HAZE dataset. In particular, DehazeDCT surpasses the second best method by an average of 0.13 dB PSNR and 0.005 SSIM, underscoring the impressive capability of our proposed method.

\Paragraph{Qualitative Comparisons}
The visual comparisons of the dehazing results from Fig.~\ref{ntire20}, ~\ref{ntire21}, and ~\ref{ntire23} demonstrate that our method achieves superior dehazing effects, capable of producing colors that are more closely aligned with the ground truth, manifesting higher clarity, greater color fidelity, and more distinct details. In contrast, other methods present distinct shortcomings. For instance, the FFA method appears to struggle with color retention, resulting in a washed-out appearance. TDN and DehazeFormer leave the image with a slightly hazy residue that obscures finer details.

\begin{table}[t]
    \setlength{\abovecaptionskip}{2mm}
    \centering
    \scalebox{1}{
    \begin{tabular}{c|cc}
    \hline
        
    \textbf{Configurations} &PSNR$\uparrow$  &SSIM$\uparrow$ \\ 
    \hline
    w/o Refinement module &21.50 &0.736 \\
    w/o Dehazing module &20.56 &0.728 \\
    only main branch &21.44 &0.730 \\    
    only frequency branch &20.07 &0.697 \\
    \hline
    \end{tabular}
    }    
\caption{The ablation result of our method on DNH-HAZE\cite{NTIRE_Dehazing_2024} dataset. Each component of our DehazeDCT contribute positively to our final dehazing performance.}
\label{tab_abla}
\end{table}
\begin{table}[t]
    \setlength{\abovecaptionskip}{2mm}
    \centering
    \scalebox{1}{
    \begin{tabular}{cccc|cc}
    \hline
    $L_1$ & $L_{Percep}$ & $L_{SSIM}$ & $L_{adv}$ & PSNR & SSIM  \\
    \hline
    \checkmark & \checkmark & \checkmark & \checkmark &21.50 &0.736  \\
    \checkmark & \checkmark & \checkmark & & 21.43 & 0.733 \\ 
    \checkmark & \checkmark & & & 21.21 & 0.725 \\
    \checkmark & & & & 21.17 & 0.720 \\
    \hline
    \end{tabular}
    }    
\caption{We conduct experiments to illustrate the rationality of loss function used for Stage I training (w/o Refinement module). Our adopted loss function help achieve optimal performance.}
\label{tab_loss}
\end{table}

\begin{table*}[!h]
    \setlength{\abovecaptionskip}{2mm}
    \centering
    \scalebox{1}{
    \begin{tabular}{c|cccc|cc}
    \hline
        
    \textbf{Team} & PSNR$\uparrow$ & SSIM$\uparrow$ & LPIPS$\downarrow$ & MOS$\uparrow$ & Average Rank & Fianl Rank$\downarrow$  \\ 
    \hline
    USTC-Dehazers & \red{\textbf{22.94}} & \red{\textbf{0.7294}} & 0.3520 & \red{\textbf{6.315}} & \red{\textbf{2.25}} & \red{\textbf{1}}\\
    \textbf{Dehazing\_R (Ours)} & \blue{22.84} & 0.7253 & 0.3466 & \blue{5.96}  & \blue{3.25} & \blue{2}\\
    Team Woof & 22.60 & \blue{0.7269} & 0.3809 & 5.79 & 4.25 & 3\\
    ITB Dehaze   & 22.32 & 0.7149 & 0.3337 & 5.705 & 4.25 &4\\
    TTWT   & 21.93 & 0.7146 & 0.3345 & 5.675 &5.25 &5 \\
    DH-AISP &21.90 & 0.7144 & 0.4017 &5.81 & 6 & 6 \\
    BU-Dehaze &21.68 & 0.7094 & \blue{0.3267} &5.22 & 6.5 & 7 \\
    RepD &21.78 & 0.7061 & 0.3328 &4.83 & 7 & 8 \\
    PSU Team &20.54 & 0.6328 & \red{\textbf{0.2678}} &5.31 & 8.75 & 9 \\
    xsourse &21.66 & 0.6955 & 0.4493 &5.28 & 9.75 & 10 \\
    \hline
    \end{tabular}
    }    
\caption{Final ranking (top 10 teams) of NTIRE 2024 Dense and Non-Homogeneous Dehazing Challenge~\cite{NTIRE_Dehazing_2024}. Our team (Dehazing\_R) achieves the \textbf{second best} performance among all submitted solutions (16 submissions in total). [Key: \textbf{\red{Best}}, \blue{Second Best}, $\uparrow (\downarrow)$: The larger (smaller) represents the better performance].}
\label{challenge_ranking}
\end{table*}


\SubSection{Ablation Study}
\label{sec_ex_abla}
To analyze the effectiveness of each component in our proposed method and justify the optimization objective utilized for training, we conduct extensive ablation experiments on DNH-HAZE Dataset. Since our method contains two separate modules and our dehazing module includes two branches, we adopt a break-down ablation to study the effectiveness of each module and each branch for dehazing.

\Paragraph {Effectiveness of Refinement Module} To study the importance of refinement module, we remove this module from our model and Tab.~\ref{tab_abla} reports the quantitative performance of remaining architecture, which still demonstrate competitive performance compared to the current SOTA in Tab.~\ref{table_quant_compar}. However, compared to our DehazeDCT, only utilizing the dehazing module suffers from obvious decrease in PSNR and SSIM, which vividly underscores the contributions of Refinement module in our proposed method. In addition, various combinations of loss functions are compared in Tab.~\ref{tab_loss}, showing that the loss function we adopted in Stage I training is reasonable and effective.

\Paragraph {Importance of Dehazing Module} In order to evaluate the importance of our proposed Dehazing module, based on our DehazeDCT, we remove the Dehazing module and directly leverage the transformer based Refinement module for dehazing and we obtain the result as Tab.~\ref{tab_abla} row 2. Compared to the result of our complete model reported in Tab.~\ref{table_quant_compar}, the significantly lower PSNR and SSIM values indicate that employing the Refinement module alone is insufficient for effective dehazing, underscoring the critical importance of the integration our Dehazing module.

\Paragraph {Contributions of DCNFormer blocks and Frequency-Aware Branch}
To further evaluate the effectiveness of each branch in the Dehaizng module, we separately adopt each branch without Refinement module to examine their contributions for dehazing. As shown in Tab.~\ref{tab_abla} row 3 and row 4, the inferior results of main branch and frequency-aware branch, compared to the complete Dehazing module, verify the beneficial contributions of each branch for dehazing and substantiate the rationality of our model's architecture. Furthermore, compared to the frequency-aware branch, our proposed DCNFormer blocks demonstrate more powerful representation learning and achieve more pleasant result.

\subsection{Performance of Our Method on NTIRE 2024 Dense and Non-Homogeneous Challenge}

The challenge results are evaluated by PSNR, SSIM, Learned Perceptual Image Patch Similarity (LPIPS)~\cite{LPIPS} and Mean Opinion Score (MOS) via a user study~\cite{NTIRE_Dehazing_2024}. The quantitative results of the top 10 teams are shown in Tab.~\ref{challenge_ranking} and our solution achieves the second best performance in total. Specifically, apart form the first solution, our method outperforms other solutions by a margin, which is quantitatively evidenced by an augmentation of 0.24 dB in PSNR and a notable elevation of 0.17 in MOS. Additionally, we provide our results of the official test set and validation set in Fig.~\ref{figure_challenge} and Fig.~\ref{fig_valid}, respectively. Obviously, it can be observed that the results generated by our model exhibit high fidelity and visual appeal. In dense hazy areas, there are no residual traces of haze and disjointed feeling with other regions. Additionally, our results feature well-defined details and vivid color, showing the superiority of our model.

\section{Conclusion}
\label{sec:conclu}
In this paper, we introduce an effective non-homogeneous \textbf{Dehaz}ing m\textbf{e}thod based on \textbf{D}eformable \textbf{C}onvolutional \textbf{T}ransformer (\textbf{DehazeDCT}). Specifically, we design a transformer-like architecture for effective dehazing with the core operator of deformable convolution v4, which offers long-range dependency and adaptive spatial aggregation capabilities and demonstrates faster convergence and forward speed. Additionally, we use a streamlined transformer grounded in Retinex theory to further improve the color and structure details. Comprehensive experiment results demonstrate the effectiveness of our proposed method. Furthermore, our method achieves outstanding performance in NTIRE 2024 Dense and Non-Homogeneous Dehazing Challenge (\textbf{second best} performance among 16 solutions).

\clearpage
{
    \small
    \bibliographystyle{ieeenat_fullname}
    \bibliography{main}
}
\end{document}